\documentclass[letterpaper]{article}
\usepackage{arxiv}
\usepackage[hyphens]{url}
\usepackage{graphicx}
\usepackage{natbib}
\usepackage{caption}
\usepackage[ruled,linesnumbered]{algorithm2e}

\usepackage{booktabs}
\usepackage{amsfonts}
\usepackage{nicefrac}
\usepackage{mathtools,bm,amssymb,xcolor}
\usepackage{subcaption,placeins}
\usepackage{adjustbox,multirow,dcolumn}
\usepackage{xstring}
\usepackage{cleveref}
\usepackage{graphicx}
\usepackage{tikz}
\usepackage{pgfplots}
\usepackage{comment}
\usepackage{etoolbox}

\newtoggle{arxiv}
\togglefalse{arxiv}

\newcommand*\samethanks[1][\value{footnote}]{%
	\footnotemark[#1]\hspace{0.4em}}

\usepackage{amsmath,amsfonts,bm}

\def\1{\bm{1}}

\DeclareMathAlphabet{\mathsfit}{\encodingdefault}{\sfdefault}{m}{sl}
\SetMathAlphabet{\mathsfit}{bold}{\encodingdefault}{\sfdefault}{bx}{n}

\makeatletter
\DeclareRobustCommand\onedot{\futurelet\@let@token\@onedot}
\def\@onedot{\ifx\@let@token.\else.\null\fi}

\newcommand{\eg}{\emph{e.g\@\onedot}}

\newcommand{\versus}{\emph{vs\@\onedot}}

\graphicspath{{figures/}}
\newcommand{\figwidth}{%
	\iftoggle{arxiv}{0.7\textwidth}{\columnwidth}%
}

\usetikzlibrary{external,arrows.meta,calc,positioning,shapes.geometric}
\usepgfplotslibrary{colorbrewer}
\pgfplotsset{compat=newest}
\newcolumntype{d}{D{.}{.}{3.2}}
\makeatletter
\newcolumntype{B}{>{\boldmath\DC@{.}{.}{3.2}}c<{\DC@end}}
\makeatother

\newcommand{\method}{\emph{Growing Harness}}

\DeclarePairedDelimiter{\parens}{\lparen}{\rparen}

\newcommand{\hpad}{\hspace*{\fill}}
\newcommand{\subplotsizeslack}{1}
\NewDocumentCommand{\subplot}{ O{1} O{} m m m }{%
	\hpad%
	\begin{subfigure}{(\textwidth / #1) * \subplotsizeslack}
		\includegraphics[width=\textwidth, #2]{#3}
		\caption{#4}\label{#5}
	\end{subfigure}%
}

\toggletrue{arxiv}

\usepackage{float}

\tikzexternaldisable

\title{Grow the Harness, Not the Context: \\
From Strategy-Free Scaffolds to Reusable Specialist Agents}
\renewcommand{\shorttitle}{Grow the Harness, Not the Context}

\author{%
  Laizhen Li\textsuperscript{1,2}\thanks{These authors contributed equally
    to this work.},
  Jiarui Li\textsuperscript{1}\samethanks,
  Juanjuan Zhao\textsuperscript{1},
  Kejiang Ye\textsuperscript{1,3},
  Ye Li\textsuperscript{1},
  Cheng-zhong Xu\textsuperscript{4},
  Xitong Gao\textsuperscript{1,3}\thanks{Corresponding author:
    \texttt{xt.gao@siat.ac.cn}.}\\[0.6em]
  \normalfont\small
  \textsuperscript{1}Shenzhen Institutes of Advanced Technology,
  Chinese Academy of Sciences\\
  \textsuperscript{2}University of Chinese Academy of Sciences\\
  \textsuperscript{3}Shenzhen University of Advanced Technology\\
  \textsuperscript{4}Institute of AI and Brain Sciences, CS Dept., University of Macau
}
\date{}

\begin{document}

\maketitle

\begin{abstract}
Large language model (LLM) agents often handle streams of related tasks,
yet standard harnesses repeatedly ask the model
to reconstruct the same control decisions
inside each task's context.
We study whether task feedback can instead turn recurring control
into reusable executable code,
while reserving LLM calls
for task-specific semantic reasoning.
We introduce \method{},
a failure-guided training paradigm
that learns the agent harness itself
from a strategy-free scaffold
that exposes fixed model and tool interfaces
but encodes no task-solving controller.
Function-level execution traces localize each failure
to a bounded code surface,
an optimizer repairs a window of failures jointly,
and a success-first held-out gate
rolls back repair sequences that harm prior capability.
Accepted edits accumulate in one shared harness,
allowing its control structure to emerge from task feedback.
Across BrowseComp-Plus and WebArena-Verified
with three deployment models from 4B to 120B parameters,
\method{} achieves the highest mean success
in five of six benchmark--model settings
and trails the best mean by 0.7 pp. in the sixth.
Relative to a Tool-Calling agent,
it reduces LLM calls by 76.0--91.8\%
and deployed-agent inference cost by 74.4--98.6\%.
On WebArena-Verified,
its success remains 44.7--45.3\% across model scales,
whereas Tool-Calling falls to 6.7\% with the 4B model.
Ablations show that trace-local edits, joint repair,
and gate-based rollback
each improve final success.
These results show that persistent program growth
can move recurring control out of model context
and into low-cost code,
yielding reusable specialist agents
that remain effective with smaller deployment models.
\end{abstract}

\section{Introduction}\label{sec:intro}

Large language model (LLM) agents solve complex tasks
by interleaving model reasoning with external tools.
In many deployments, however,
an agent does not face isolated tasks.
It repeatedly handles instances from the same task family,
using the same model and tool interfaces.
Although goals and observations change across instances,
the surrounding control often recurs:
the agent must refine queries,
filter observations,
verify progress,
recover from errors,
and decide when to stop.
Agent harnesses,
the executable code that orchestrates model and tool calls,
either specify such behavior in advance
or delegate it to LLMs
\cite{wang2024survey,yao2023react}.
Repeated delegation preserves flexibility,
but every decision incurs another inference call,
grows context,
and adds prompts, observations, and outputs
to a growing execution history.
\begin{figure}[t]
    \centering
    \includegraphics[
        width=\figwidth, trim=0 5pt 0 5pt,
    ]{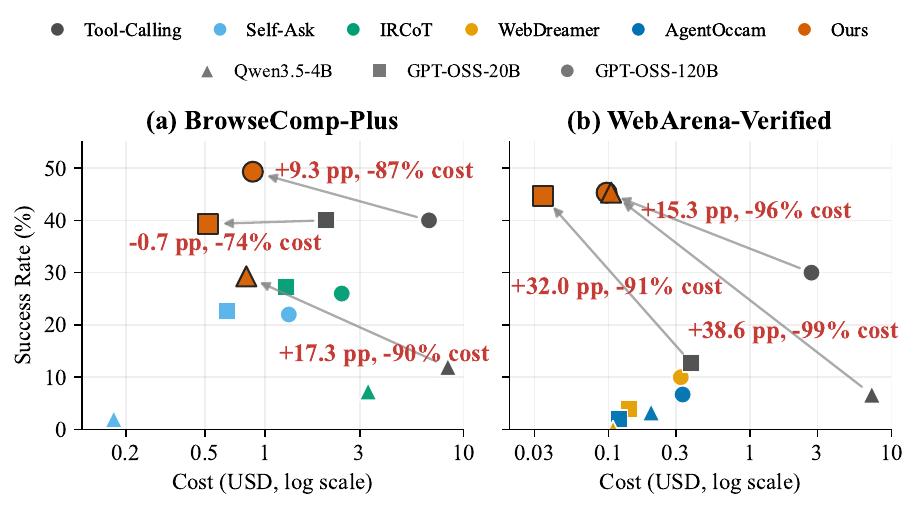}
    \caption{%
        Success--cost trade-offs on BrowseComp-Plus and WebArena-Verified.
        Color and marker shape encode the agent and deployment model,
        respectively; each point is a mean over three independent runs,
        with 95\% confidence intervals omitted for clarity and reported in
        \Cref{tab:main-results}.}
    \label{fig:cost-success}
\end{figure}

Repeated deployment within a task family
therefore creates an opportunity for \emph{harness growth}.
Here,
harness growth means using task feedback
to turn recurring control into persistent executable behavior.
Across tasks drawn from a common distribution,
the harness can accumulate code
that later executions reuse
instead of reconstructing the same behavior
through online inference.
Prior agents reuse experience
through retrieved workflows,
executable skills,
or synthesized tools
\cite{wang2024awm,wang2023voyager,cai2023latm,yuan2023craft}.
These approaches show that behavior learned from earlier tasks
can benefit later executions.
Harness growth uses the agent program itself
as the persistent artifact:
behavior acquired from earlier failures
runs directly at low marginal cost,
while the LLM remains available
for task-dependent semantic reasoning.
This leads to our central question:
how can a scaffold with no predefined controller grow from task feedback
so that it stops asking the LLM
to reconstruct behavior
that it can execute in code?
Our thesis is that such growth
can manage LLM context structurally:
the harness acquires recurring control
as persistent executable code,
reserving model context
for task-specific evidence and semantic reasoning.
This design is especially useful
for small LLMs deployed on mobile
and other resource-constrained devices:
moving recurring control from inference into code
offers a path to capable agents
when model scale and online inference are constrained.

A harness grown from task feedback
should meet three requirements.
\textbf{(a) reuse:}
behavior acquired from training tasks
should help unseen tasks from the same distribution,
rather than encode instance-specific solutions.
\textbf{(b) context-efficient execution:}
the harness should reduce repeated model calls
and avoid placing routine control into a growing model context,
while preserving LLM reasoning where semantics matter.
Long histories increase input cost
and can make relevant evidence harder to use
\cite{yao2023react,liu2024lostmiddle}.
\textbf{(c) low prior commitment:}
harness growth
should not require a complete controller at initialization.
Recent methods optimize existing harnesses
or synthesize code within prescribed interfaces and structures
\cite{lou2026autoharness,lee2026metaharness}.
We call this starting point a \emph{strategy-free scaffold}:
it exposes the required task, model, and tool interfaces
but encodes no task-solving controller,
such as a ReAct-style tool-calling loop.
This allows the executable structure itself
to adapt to task feedback.

Three technical challenges emerge.
\textbf{(a) What should become code,
and what should remain an LLM call?}
Executable code is cheap to reuse,
but brittle rules cannot replace semantic judgment.
The learner must identify deterministic,
structured behavior that transfers across tasks,
while reserving model calls
for interpretation, synthesis, and other open-ended decisions.
\textbf{(b) How can an optimizer improve a growing program
without processing the whole program at every step?}
As the harness accumulates capabilities,
whole-program optimization
makes the code context and search space grow with it.
A task-level outcome identifies whether an execution succeeded,
but not which function or control decision caused a failure.
Harness growth therefore requires execution evidence
that links each failure to a bounded local code surface,
so per-step optimization depends on the active execution slice
rather than the total program size.
\textbf{{(c) How can growth avoid brittleness and regression?}}
An edit that repairs one task
may encode an instance-specific rule
or damage behavior acquired earlier.
This risk is acute during continual harness optimization,
where successive local gains may fail to compound
\cite{wang2026compound}.
The learner must favor behavior shared across failures
and test each update independently
before adding it to the deployed harness.

We introduce \textbf{\method{}},
a failure-guided training paradigm
that naturally grows a cost-efficient agent harness
from a strategy-free executable scaffold.
The scaffold exposes the task entry point
and fixed LLM and tool interfaces,
but encodes no task-solving controller.
Training runs the current harness on a task stream
and maintains a bounded window of failures.
For each failed execution,
a function-level execution DAG records
the code paths, model calls, tool calls, and errors involved.
An offline optimizer uses these traces
to synthesize a complete executable candidate,
while trace scope and an edit budget
limit changes to implicated functions
and a small number of new helpers.
We direct the optimizer
to implement deterministic and reusable operations in code,
while retaining LLM calls
for task-dependent semantic reasoning.
Candidates are tested on the active failures,
and a success-first held-out gate
rejects repair sequences
that reduce aggregate gate-set success.
Accepted edits update the shared harness,
so executable behavior learned from one group of failures
can serve later tasks.
Over successive rounds,
the scaffold grows into a code-first, LLM-assisted agent
without committing to a predefined controller.
\begin{figure*}[t]
    \centering
    \includegraphics[width=\textwidth, trim=0 10pt 0 0]{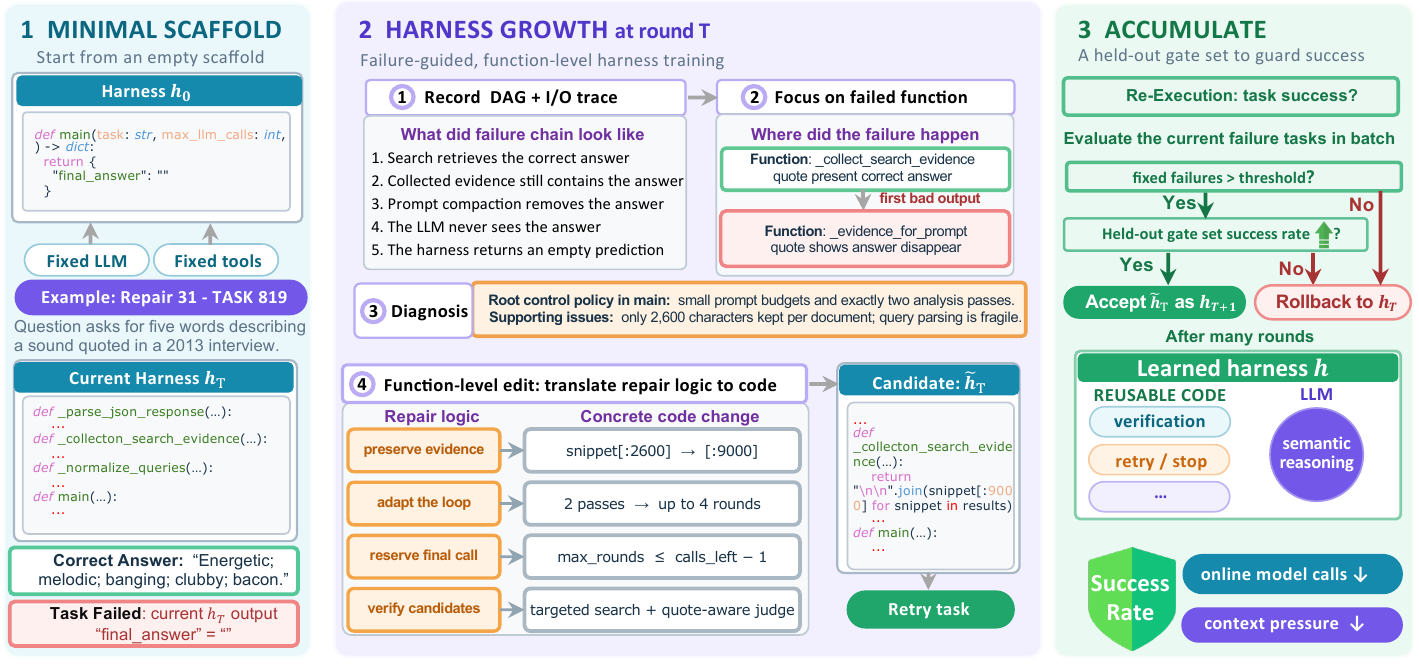}
    \caption{%
Overview of the \method{}. From a strategy-free scaffold \(h_0\), each round traces a failed execution, locates the faulty functions, diagnoses the underlying control-policy failure, and repair function-level code. Accept candidate \emph{iff} it fixes enough failures and preserves held-out success.
    }
    \label{fig:natural-growth}
\end{figure*}

\Cref{fig:cost-success}
previews the resulting success--cost trade-off.
Across BrowseComp-Plus and WebArena-Verified,
the learned harnesses occupy the high-success, low-cost region,
with the largest separation from inference-intensive agents
on WebArena-Verified and with the 4B deployment model,
highlighting the potential of harness growth
for small LLMs in mobile and other constrained deployments.

Our contributions are:
\begin{itemize}
    \item We formulate agent learning
    as reusable program growth from task feedback,
    starting from a strategy-free scaffold
    under fixed LLM and tool interfaces.

    \item We introduce trace-local program growth,
    which combines multi-failure training
    with function-level trace-scoped edits
    so the global harness can grow
    while each optimizer step remains focused
    on a bounded execution slice.
    Held-out gate rollback guards
    against brittle or regressive updates.

    \item Across BrowseComp-Plus and WebArena-Verified
    and three deployment-model scales,
    \method{} achieves the highest mean success rate
    in five of six settings
    while reducing LLM calls by 76.0--91.8\%
    and online cost by 74.4--98.6\%
    relative to Tool-Calling.
    Ablations show complementary gains
    from function-level guidance,
    the failure-window curriculum,
    and gate-based rollback.
\end{itemize}

\section{Related Work}\label{sec:related}

\paragraph{Inference-time control in tool-using agents.}
LLM agents rely on control structures around the model
to organize reasoning, tool use, observation handling, and stopping.
ReAct interleaves reasoning and acting,
Self-Ask decomposes questions into follow-up queries,
and IRCoT alternates retrieval with multi-step reasoning
\cite{yao2023react,press-etal-2023-measuring,
trivedi-etal-2023-interleaving}.
Reflexion further carries feedback from earlier attempts
into subsequent executions \cite{Reflexion}.
These methods establish that the control surrounding an LLM
is a substantive part of agent behavior.
Their loop structures are nonetheless specified before deployment,
while many control decisions are represented in textual histories
and recomputed by the model as each trajectory unfolds.
Harness growth asks a different question:
how can task feedback change the executable control itself,
with the LLM and tools fixed,
so later tasks reuse code rather than repeat inference?

\paragraph{Persistent experience across tasks.}
Experiential agents such as ExpeL and Agent Workflow Memory
store natural-language insights, examples, or workflows
for later retrieval
\cite{zhao2023expel,wang2024awm}.
Voyager, LATM, and CRAFT instead distill experience
into executable skills or tools
that can be reused across tasks
\cite{wang2023voyager,cai2023latm,yuan2023craft}.
Both lines of work show that experience
can be amortized across related tasks.
Textual artifacts,
however,
must be retrieved and interpreted within the model context,
while skills package individual behaviors
behind an existing agent structure.
Our persistent artifact is the shared harness itself:
accepted repairs become ordinary program paths
that orchestrate model and tool use for all later tasks.
The result is a growing controller,
rather than a collection of experiences
that the agent must retrieve and reinterpret online.

\paragraph{Optimizing language-model programs.}
A broad line of work improves LLM systems
without updating model weights.
APE, OPRO, and related methods
optimize natural-language instructions
\cite{zhou2023ape,yang2023opro,pryzant2023protegi},
while DSPy, MIPRO, TextGrad, and GEPA
optimize prompts and demonstrations
across multi-stage language-model programs
\cite{khattab2023dspy,opsahlong2024mipro,
yuksekgonul2024textgrad,agrawal2025gepa}.
AFlow and ADAS search over code-represented workflows
or broader agentic system designs
using task feedback
\cite{zhang2024aflow,hu2024adas}.
These methods establish prompts, modules,
and workflow structure as learnable objects.
Our focus is continual program growth
from a strategy-free scaffold:
failed executions identify missing behavior,
trace-local edits add that behavior to one shared harness,
and accepted updates accumulate across a task stream.

\paragraph{Harness optimization.}
Recent work directly optimizes the code and configuration
surrounding an LLM application.
AutoHarness synthesizes code constraints or complete policies
from environment feedback,
and Meta-Harness searches over model-harness code
using source code, candidate scores, and execution traces
\cite{lou2026autoharness,lee2026metaharness}.
VeRO provides versioning, structured observations,
and evaluation infrastructure
for agents that optimize other agents
\cite{ursekar2026vero}.
Other methods learn from retrospective or retrieved experience,
or adapt harnesses so smaller models
can approach the performance of stronger models
\cite{pan2026rho,huang2026memoharness,yang2026betterharness}.
Recent evidence also shows
that successive harness updates can lose earlier gains
without explicit regression control
\cite{wang2026compound}.
Our distinction is not code synthesis alone,
but open-ended global harness growth
through local, failure-conditioned optimization,
with training beginning from a strategy-free scaffold
under fixed LLM and tool interfaces.

\paragraph{Cost-efficient LLM and agent execution.}
LLMLingua compresses prompts,
whereas FrugalGPT and RouteLLM use cascades or learned routing
to invoke expensive models selectively
\cite{jiang2023llmlingua,chen2023frugalgpt,ong2025routellm}.
AgentOccam and WebDreamer reduce planning or interaction overhead
through more efficient agent architectures
\cite{yang2025agentoccam,gu2025webdreamer},
and agent evaluations increasingly report monetary cost
alongside task success \cite{kapoor2025hal}.
These approaches make calls shorter or cheaper,
select a cheaper call,
or improve a fixed inference-time strategy.
Our approach instead manages context structurally:
it grows executable code
so recurring control need not be regenerated
or carried through the model context on every task.
The deployed LLM and tools remain fixed,
and model context is reserved
for task-specific evidence and semantic reasoning.

\section{Method}\label{sec:method}

\subsection{Problem Formulation}\label{sec:method:formulation}

An agent workflow
can be represented as the composition of an executable harness,
a language model backend, and a set of external tools.
Let \(M\) denote a fixed LLM backend,
let \(T\) denote a fixed collection of tool interfaces,
and let \(h\in\mathcal{H}\) denote executable harness code.
Given a task \(x\),
the deployed agent harness is \( \textstyle h(x;M,T) \),
which specifies how the agent invokes \(M\) and \(T\),
maintains intermediate state,
processes observations, recovers from errors,
and decides when to terminate.
In our paradigm,
\(M\) and \(T\) remain fixed;
the harness \(h\) is the only learnable object.

For a task set \(D\),
we denote the success rate of a harness \(h\)
as \(\mathrm{SR}_{D} (h)\).
We partition the available tasks
into a training set \(D_{\mathrm{tr}}\),
a gate validation set \(G\),
and a final evaluation set \(D_{\mathrm{te}}\).
The training set supplies execution experience,
the gate set controls checkpoint acceptance,
and the final set is used only for reporting.
The optimizer never observes outcomes or traces
from \(D_{\mathrm{te}} \).

Training starts from a minimal executable scaffold \(h_0\).
The scaffold exposes the required task entry point
and fixed LLM and tool interfaces,
but contains no complete hand-designed agent strategy.
Because it retains generic access to \(M\) and \(T\),
an LLM-mediated agent remains representable
within the harness hypothesis class.
Training can therefore grow executable control
from a weak initialization
without committing to a predefined controller
such as native tool-calling loop or Self-Ask.

\subsection{Harness Execution and Function-Level Traces}\label{sec:method:traces}

For each task \(x_i\),
the runtime records an execution graph
\( \mathcal{G}_i= (\mathcal{V}_i, \mathcal{E}_i) \).
Nodes represent harness-function invocations,
LLM and tool calls, returns, or runtime errors,
and record the operation, inputs, outputs or errors, parent, and duration.
An edge \((u,v)\)
indicates that \(v\) executes within the dynamic scope of \(u\).
The task trace is
\begin{equation}\textstyle
    \tau_i = \left(x_i, \mathcal{G}_i, y_i,o_i,u_i\right),
\end{equation}
where \(y_i\) is the output,
\(o_i=o(h,x_i) \in\{0,1\}\)
is the benchmark outcome,
and \(u_i\) contains the token, cost, and runtime measurements.
To localize a failure to participating code,
we extract
\begin{equation}\textstyle
    \mathcal{F}(\tau_i) = \left\{
        f : f \text{ is a harness function invoked in } \mathcal{G}_i
    \right\}.
\end{equation}
This set later constrains
which existing functions the optimizer may modify.
\begin{algorithm}[!t]\label{sec:method:algorithm}
	\caption{Failure-guided harness growth.}\label{alg:harness-growth}
	\iftoggle{arxiv}{}{\small\SetAlgoNoEnd}
	\DontPrintSemicolon

	\KwIn{Training set \(D_{\mathrm{tr}}\), gate set \(G\),
		fixed model \(M\), tools \(T\), optimizer \(O\),
		initial harness \(h_0\), window size \(K\),
		repair threshold \(Q\), attempt limit \(R_{\max}\),
		and edit budget \(L\)}
	\KwOut{Learned harness \(h^\star\)}

	\(t\gets0;\quad W_t\gets\varnothing\)\;

	\While{unseen tasks remain or \(W_t\neq\varnothing\)}{
		Run \(h_t\) on unseen tasks and retain failures and traces
		until \(|W_t|=K\) or the training stream ends\;

		\If{\(W_t=\varnothing\)}{\textbf{break}}

		Using \(h_t\) and the traces in \(W_t\), \(O\) produces
		a valid trace-scoped candidate \(\widetilde h_t\) satisfying
		\(d_{\mathrm{fun}}(h_t,\widetilde h_t)\leq L\)\;

		Re-execute \(\widetilde h_t\) on every task in \(W_t\)\;
		\(P_t\gets
		\{x_i\in W_t:o(\widetilde h_t,x_i)=1\}\)\;
		\(U_t\gets W_t\setminus P_t\)\;

		\If{\(|P_t|<Q\)}{
			Discard \(\widetilde h_t\) and retain \(h_t\)
			\tcp*[r]{rollback: insufficient repairs}
			Increment attempt counts in \(W_t\) and retire tasks
			reaching \(R_{\max}\)\;
			\textbf{continue}\;
		}

		\If{\(\mathrm{SR}_G(\widetilde h_t)
			<\mathrm{SR}_G(h_t)\)}{
			Discard \(\widetilde h_t\) and retain \(h_t\)
			\tcp*[r]{rollback: gate regression}
			Increment attempt counts in \(W_t\) and retire tasks
			reaching \(R_{\max}\)\;
			\textbf{continue}\;
		}

		Accept the candidate:
		\(h_{t+1}\gets\widetilde h_t\)\;
		Set \(W_{t+1}\) to the unresolved tasks \(U_t\), with fresh
		traces and incremented attempt counts; retire tasks reaching
		\(R_{\max}\)\;
		\(t\gets t+1\)\;
	}

	\(h^\star\gets h_t\); \quad \textbf{return} \(h^\star\)\;
\end{algorithm}

\subsection{Failure-Window Curriculum}\label{sec:method:window}

Rather than repeatedly optimizing tasks already solved,
we maintain a bounded window \(W_t\)
of current failures at round \(t\).
For the ordered training sequence
\( D_{\mathrm{tr}}= (x_1,\ldots,x_N) \),
each entry stores a task,
its latest trace,
and its repair-attempt count
\( w_i= (x_i,\tau_i,a_i) \).
The current harness fills the window
with failures from unseen tasks,
up to capacity \(K\);
direct successes leave the curriculum.
The optimizer jointly repairs all window failures
to encourage reusable behavior.
After each repair,
solved tasks \(P_t\) leave.
Unresolved tasks \(U_t=W_t \setminus P_t\)
retain fresh traces and incremented attempt counts;
tasks whose updated count reaches \(R_{\max}\) are retired.
New failures \(F_t\) then refill the window.
Ignoring the stored trace metadata,
its task membership evolves
as
\begin{equation}\textstyle
    W_{t+1} = \left\{
        x_i\in U_t: a_i<R_{\max}
    \right\} \cup F_t.
\end{equation}
Here,
\(a_i\) is the updated attempt count.
Before optimizing a full window,
we checkpoint the complete state;
if every task in that window exhausts its budget
without a repair,
we restore the checkpoint.
This prevents an ineffective repair sequence from persisting
when no single edit triggers an immediate gate regression.

\subsection{Trace-Guided Function-Level Optimization}
\label{sec:method:function}

Given \(h_t\) and \(W_t\),
an offline optimizer receives
\begin{equation}\textstyle
    Z_t = \left( h_t, \{ \tau_i:x_i\in W_t \}, E \right),
\end{equation}
where \(E\) contains offline diagnostic artifacts
such as evaluator feedback, retrieved-document identifiers,
evidence, action statistics, and errors.
These artifacts are unavailable to the deployed harness.
An optimizer model \(M_{\mathrm{opt}} \),
which may differ from \(M\),
produces a complete executable candidate
\begin{equation}\textstyle
    \widetilde{h}_{t+1} = O\left( Z_t;M_{\mathrm{opt}} \right).
\end{equation}

To connect failures to edits and bound the search space,
the optimizer may modify only the entry function
and functions invoked by a supplied failure trace:
\begin{equation}\textstyle
    \mathcal{A}_t =
        \{\texttt{main}\} \cup \bigcup_{x_i\in W_t} \mathcal{F}(\tau_i).
\end{equation}
It may also add reusable helpers,
but must leave untraced existing functions unchanged.
If \(d_{\mathrm{fun}}\) counts modified and newly introduced functions,
we require
\( d_{\mathrm{fun}} (h_t, \widetilde{h}_{t+1}) \leq L \)
for edit budget \(L\).
The optimizer
must preserve function signatures and runtime interfaces,
may not delete existing functions,
and may not encode task identifiers, expected answers, or fixed solutions.

We direct code toward deterministic, reusable control,
\eg{}, parsing, validation, state updates,
conditional query refinement, error recovery, and stopping,
while retaining LLM calls for semantic interpretation,
synthesis, fuzzy comparison, and answer generation.
Before benchmark evaluation,
each candidate must parse and compile,
expose the required entry point,
preserve the fixed model and tool interfaces,
and satisfy the trace-scope
and edit-budget constraints.
Invalid candidates are rejected.

\subsection{Candidate Evaluation and Window Update}\label{sec:method:candidate}

A valid candidate is re-executed on the entire active window.
The solved and unresolved subsets are
\begin{equation}\textstyle
    P_t
    =
    \left\{x_i\in W_t:o(\widetilde{h}_{t+1},x_i)=1\right\},
    \quad
    U_t=W_t\setminus P_t.
\end{equation}
The candidate becomes the provisional harness,
\(h_{t+1}\gets
\widetilde{h}_{t+1}\);
\(P_t\) leaves the window,
whereas \(U_t\) receives incremented attempt counts and fresh traces.
Because each repair is a complete harness used on subsequent tasks,
helpers and control logic accumulate across rounds
rather than remaining task-specific patches.
If synthesis or validation fails,
optimization is retried without changing \(h_t\);
exhausting the optimizer's internal retry limit
restores the latest accepted checkpoint and stops training.
The harness, window, task cursor, counters, and checkpoint metadata
are serialized for resumption.

\subsection{Held-Out Gate Validation and Rollback}\label{sec:method:gate}

Repairs may solve the active window
while damaging earlier behavior.
We therefore evaluate \(h_0\)
on held-out gate set \(G\),
then reevaluate the current harness
after every \(Q\) tasks solved through repair.
Relative to the latest accepted checkpoint
\(h^{\mathrm{gate}}\),
we accept \(h_t\) \emph{iff}
\(
    \mathrm{SR}_G(h_t)
    \geq
    \mathrm{SR}_G(h^{\mathrm{gate}})
\).
Once accepted,
the complete current state
becomes the new checkpoint \( h^{\mathrm{gate}}\leftarrow h_t \).
Otherwise,
the entire repair sequence
since the previous gate is rolled back.
Rollback restores not only code,
but also the task cursor,
failure window, counters, and training records,
yielding the same state from which the rejected sequence began.

After the training stream and window are exhausted,
any changes made since the last checkpoint undergo one final gate evaluation.
The current harness is selected only if its gate success is no lower
than that of \(h^{\mathrm{gate}} \);
otherwise, the latest accepted checkpoint is restored.
The resulting \(h^\star\) is evaluated once on \(D_{\mathrm{te}}\).
The rule is success-first:
lower online cost cannot compensate for lower gate success,
although cost is recorded for all rollouts
to compare success-preserving harnesses.

In summary,
failures localize supervision for bounded program synthesis,
repairs accumulate reusable control in one executable harness,
and transactional gate rollback protects previously acquired capability
while the fixed LLM continues to provide task-dependent semantic reasoning.

\section{Experiments}\label{sec:results}

\subsection{Experimental Setup}\label{sec:results:setup}

\noindent\textbf{Datasets.}
We evaluate \method{}
on \textbf{BrowseComp-Plus} \citep{chen2025BrowseCompPlus},
a controlled benchmark for deep-search agents
under a fixed, human-verified retrieval corpus,
and \textbf{WebArena-Verified} \citep{hattami2025webarena},
an audited benchmark for reproducible evaluation of multi-step web agents
with corrected tasks and deterministic evaluators.
The two benchmarks cover complementary settings:
open-domain retrieval and evidence synthesis in BrowseComp-Plus,
and multi-step interaction in WebArena-Verified.
For each benchmark,
we use 200 training tasks,
50 held-out gate tasks,
and 50 final-evaluation tasks.

\noindent\textbf{Models.}
We evaluate the trained harnesses
on three deployment LLMs: gpt-oss-120b and gpt-oss-20b
\citep{openai2025gptoss120bgptoss20bmodel},
and Qwen3.5-4B \citep{qwen3.5}.

\noindent\textbf{Baselines.}
We compare the trained harnesses
with representative inference-time agent strategies
adapted to each benchmark.
\textbf{Tool-Calling}
is a fixed, non-learning agent
that repeatedly selects an environment tool
based on the task context and interaction history.
Its interleaved tool-calling loop
follows a reasoning-and-acting execution pattern \citep{yao2023react}.
For BrowseComp-Plus,
we adapt \textbf{IRCoT},
which interleaves chain-of-thought reasoning
with retrieval \citep{trivedi-etal-2023-interleaving};
\textbf{Self-Ask},
which decomposes a question into explicit follow-up questions
that can be answered through search \citep{press-etal-2023-measuring}.
For WebArena-Verified,
we adapt \textbf{WebDreamer},
which uses an LLM as a world model
to predict and evaluate the outcomes
of candidate browser actions \citep{gu2025webdreamer},
and \textbf{AgentOccam},
which aligns the browser observation and action spaces
with the capabilities of the underlying LLM \citep{yang2025agentoccam}.
Within each benchmark,
all methods share deployment LLMs,
task splits, tools, and evaluation protocol;
complete baseline configurations and our hyperparameters
are provided in \iftoggle{arxiv}{the appendix}{the supplementary material}.

\begin{table*}[t]
\centering\begin{adjustbox}{max width=\textwidth}
\begin{tabular}{cllcccccc}
\toprule
& \textbf{Model}
& \textbf{Agent}
& \textbf{SR (\%) \(\uparrow\)}
& \textbf{\#Calls \(\downarrow\)}
& \textbf{Input (k) \(\downarrow\)}
& \textbf{Output (k) \(\downarrow\)}
& \textbf{Time (s) \(\downarrow\)}
& \textbf{Cost (\$) \(\downarrow\)} \\
\midrule
\multirow{12}{*}{\rotatebox[origin=c]{90}{\textbf{BrowseComp-Plus}}} & \multirow{4}{*}{gpt-oss-120b} & Tool-Calling & $40.0 \pm 11.6$ & $32.7 \pm 3.4$ & $1599.1 \pm 238.0$ & $12.5 \pm 1.9$ & $241.1 \pm 39.4$ & $6.70 \pm 0.96$ \\
& & Self-Ask & $22.0 \pm 10.6$ & $37.5 \pm 4.8$ & $168.0 \pm 24.4$ & $15.2 \pm 2.0$ & $522.3 \pm 76.8$ & $1.32 \pm 0.18$ \\
& & IRCoT & $26.0 \pm 11.0$ & $36.4 \pm 5.3$ & $312.1 \pm 48.2$ & $10.0 \pm 1.7$ & $521.2 \pm 89.2$ & $2.44 \pm 0.37$ \\
& & \textbf{Ours} & $\mathbf{49.3 \pm 11.7}$ & $\mathbf{6.0 \pm 0.0}$ & $\mathbf{86.9 \pm 2.1}$ & $\mathbf{7.5 \pm 0.7}$ & $\mathbf{176.9 \pm 11.2}$ & $\mathbf{0.87 \pm 0.03}$ \\
\cmidrule(lr){2-9}
& \multirow{4}{*}{gpt-oss-20b} & Tool-Calling & $\mathbf{40.0 \pm 11.7}$ & $24.8 \pm 3.3$ & $952.8 \pm 196.0$ & $\mathbf{8.3 \pm 1.5}$ & $407.0 \pm 71.3$ & $2.03 \pm 0.40$ \\
& & Self-Ask & $22.7 \pm 9.1$ & $34.5 \pm 4.2$ & $131.1 \pm 19.0$ & $19.8 \pm 2.5$ & $399.6 \pm 52.0$ & $0.65 \pm 0.08$ \\
& & IRCoT & $27.3 \pm 11.2$ & $32.2 \pm 5.2$ & $275.3 \pm 47.4$ & $21.3 \pm 4.9$ & $415.4 \pm 89.6$ & $1.28 \pm 0.21$ \\
& & \textbf{Ours} & $39.3 \pm 12.6$ & $\mathbf{6.0 \pm 0.1}$ & $\mathbf{82.9 \pm 2.5}$ & $14.3 \pm 1.7$ & $\mathbf{351.8 \pm 73.9}$ & $\mathbf{0.52 \pm 0.03}$ \\
\cmidrule(lr){2-9}
& \multirow{4}{*}{qwen3.5-4b} & Tool-Calling & $12.0 \pm 7.1$ & $29.6 \pm 3.3$ & $1569.5 \pm 263.1$ & $3.3 \pm 0.5$ & $1608.0 \pm 84.9$ & $8.33 \pm 1.36$ \\
& & Self-Ask & $2.0 \pm 2.2$ & $6.7 \pm 1.9$ & $\mathbf{21.1 \pm 8.8}$ & $\mathbf{1.1 \pm 0.3}$ & $\mathbf{262.9 \pm 85.8}$ & $\mathbf{0.17 \pm 0.06}$ \\
& & IRCoT & $7.3 \pm 7.0$ & $46.4 \pm 3.1$ & $369.0 \pm 33.0$ & $3.6 \pm 0.5$ & $529.4 \pm 84.2$ & $3.31 \pm 0.29$ \\
& & \textbf{Ours} & $\mathbf{29.3 \pm 11.8}$ & $\mathbf{5.5 \pm 0.2}$ & $81.5 \pm 4.9$ & $2.6 \pm 0.3$ & $390.2 \pm 73.7$ & $0.81 \pm 0.05$ \\
\midrule
\multirow{12}{*}{\rotatebox[origin=c]{90}{\textbf{WebArena-Verified}}} & \multirow{4}{*}{gpt-oss-120b} & Tool-Calling & $30.0 \pm 11.4$ & $26.3 \pm 4.1$ & $624.1 \pm 158.4$ & $6.9 \pm 1.5$ & $667.5 \pm 122.7$ & $2.72 \pm 0.66$ \\
& & WebDreamer & $10.0 \pm 7.0$ & $10.7 \pm 1.3$ & $19.1 \pm 2.8$ & $6.3 \pm 0.9$ & $\mathbf{199.0 \pm 36.4}$ & $0.32 \pm 0.05$ \\
& & AgentOccam & $6.7 \pm 6.0$ & $9.7 \pm 1.2$ & $20.8 \pm 3.2$ & $6.2 \pm 0.8$ & $247.7 \pm 75.1$ & $0.33 \pm 0.05$ \\
& & \textbf{Ours} & $\mathbf{45.3 \pm 13.4}$ & $\mathbf{3.8 \pm 2.6}$ & $\mathbf{8.0 \pm 6.0}$ & $\mathbf{1.6 \pm 1.2}$ & $299.9 \pm 119.1$ & $\mathbf{0.10 \pm 0.07}$ \\
\cmidrule(lr){2-9}
& \multirow{4}{*}{gpt-oss-20b} & Tool-Calling & $12.7 \pm 7.3$ & $22.3 \pm 2.7$ & $158.5 \pm 25.1$ & $3.1 \pm 0.4$ & $461.4 \pm 62.3$ & $0.39 \pm 0.05$ \\
& & WebDreamer & $4.0 \pm 3.9$ & $7.7 \pm 1.1$ & $13.3 \pm 2.1$ & $6.1 \pm 1.0$ & $174.0 \pm 30.7$ & $0.14 \pm 0.02$ \\
& & AgentOccam & $2.0 \pm 2.8$ & $6.2 \pm 0.9$ & $12.5 \pm 2.5$ & $4.9 \pm 0.8$ & $\mathbf{116.6 \pm 27.5}$ & $0.12 \pm 0.02$ \\
& & \textbf{Ours} & $\mathbf{44.7 \pm 13.4}$ & $\mathbf{1.8 \pm 1.0}$ & $\mathbf{3.5 \pm 2.3}$ & $\mathbf{1.6 \pm 1.1}$ & $250.1 \pm 100.5$ & $\mathbf{0.03 \pm 0.02}$ \\
\cmidrule(lr){2-9}
& \multirow{4}{*}{qwen3.5-4b} & Tool-Calling & $6.7 \pm 4.9$ & $37.8 \pm 2.7$ & $1355.6 \pm 161.7$ & $4.4 \pm 0.5$ & $1312.6 \pm 100.7$ & $7.26 \pm 0.85$ \\
& & WebDreamer & $0.0 \pm 0.0$ & $\mathbf{4.9 \pm 0.7}$ & $\mathbf{10.0 \pm 1.5}$ & $1.2 \pm 0.2$ & $\mathbf{174.9 \pm 30.1}$ & $0.11 \pm 0.02$ \\
& & AgentOccam & $3.3 \pm 4.6$ & $7.0 \pm 0.8$ & $17.5 \pm 3.4$ & $3.2 \pm 0.5$ & $490.3 \pm 78.1$ & $0.20 \pm 0.04$ \\
& & \textbf{Ours} & $\mathbf{45.3 \pm 13.4}$ & $5.4 \pm 3.8$ & $13.7 \pm 10.1$ & $\mathbf{0.4 \pm 0.5}$ & $387.5 \pm 148.6$ & $\mathbf{0.10 \pm 0.08}$ \\
\bottomrule
\end{tabular}
\end{adjustbox}
\caption{Main results on BrowseComp-Plus and WebArena-Verified.
Entries report means over three independent evaluation runs; \(\pm\)
values denote normal-approximation 95\% confidence-interval half-widths
estimated by task-level cluster bootstrap.
SR is the success rate, \#Calls is the number of deployed-agent LLM
calls, and Input/Output are the input/output token counts in thousands.
Calls, token counts, and time are per-task averages. Cost denotes the
mean total deployed-agent LLM cost of one 50-task evaluation run in
U.S. dollars. Best mean values within each benchmark--model block
are shown in bold.}
\label{tab:main-results}
\label{tab:webarena-main-results}
\end{table*}

\noindent\textbf{Evaluation Metrics.}
For each method--model configuration,
we conduct \(R=3\) independent evaluation runs.
In each run,
every final-evaluation task receives one single-attempt agent rollout.

Our primary task metric is \textbf{Success Rate},
defined as the percentage of successful task--run pairs:
\begin{equation}\textstyle
    \widehat{\mathrm{SR}} =
    \frac{1}{R|\mathcal{D}|}
    \sum_{r=1}^{R}
    \sum_{x_i \in \mathcal{D}}
    \mathbb{I}[\mathrm{Eval}
        (x_i,
    \hat{a}_{ir})=1
    ],
\end{equation}
where \(\mathcal{D}\) is the final-evaluation set
and \(\hat{a}_{ir}\) is the answer produced for task \(x_i\) in run \(r\).
We average results over all three runs
and do not select the best of multiple attempts.
BrowseComp-Plus uses an LLM-based equivalence judge,
whereas WebArena-Verified uses the benchmark evaluator.

For efficiency,
we report \textbf{LLM Calls},
\textbf{Input Tokens},
\textbf{Output Tokens},
\textbf{Time},
and \textbf{Total Online Cost}.
Calls, token counts, and time are averaged over all task-run pairs.
Token counts are reported in thousands per task,
and Input Tokens include both uncached and cache-read tokens.
Total Online Cost includes deployed-agent LLM calls only,
excluding evaluator and offline-optimizer calls;
it is computed separately for each run
and then averaged across the three runs.
For a run,
we compute
\begin{equation}\textstyle
    \mathrm{Cost} = \sum_{t=1}^{N_{\mathrm{req}}} \parens{
            u_t \pi_{\mathrm{in}} +c_t\pi_{\mathrm{cache}} +v_t\pi_{\mathrm{out}}
        } / {10^6},
\end{equation}
where \(u_t\), \(c_t\), and \(v_t\)
are the uncached-input, cache-read,
and output tokens for request \(t\),
respectively.
The provider-listed prices used for cost accounting
are reported in \iftoggle{arxiv}{the appendix}{the supplementary material}.
We estimate cache-read tokens
using the longest prefix
shared with an earlier request
from the same task
and do not assume cache reuse across tasks.

To quantify uncertainty,
we use a task-level cluster bootstrap
with 10,000 replicates
and a fixed random seed of 42.
Each replicate samples
\(|\mathcal{D}|\) tasks with replacement
while retaining all three runs for each sampled task.
We report every metric
as \( \text{mean} \pm 1.96\, \mathrm{SE}_{\mathrm{bootstrap}} \);
thus, each \(\pm\) value
is a normal-approximation 95\% confidence-interval half-width,
not a standard deviation.

\subsection{Main Results}\label{sec:results:main}

\noindent\textbf{\method{} transfers well to unseen tasks.}
Across two distinct task families,
\method{} achieves the highest mean success rate
in five of the six benchmark--model settings
and falls only 0.7 pp. short
of the highest mean in the remaining setting
(\Cref{tab:main-results}).
The breadth of this result suggests
that the learned programs capture behavior shared within each family,
rather than repairs specific to the training failures
that produced them.

\noindent\textbf{Replaces inference with code: 76--92\% fewer LLM calls at 74--99\% lower cost.}
Relative to Tool-Calling,
\method{} reduces LLM calls by 76.0--91.8\%
and online cost by 74.4--98.6\%
across the six settings.
Despite using fewer calls,
it improves mean success over Tool-Calling in five settings
and remains within 0.7 pp. in the sixth.
The resulting success--cost frontier in \Cref{fig:cost-success}
supports the central premise of harness growth:
recurring control can run as reusable code at low marginal cost,
without replacing the LLM's task-specific semantic reasoning.

\noindent\textbf{The learned harness sustains success across model scales.}
On WebArena-Verified,
its mean success remains within a narrow 44.7--45.3\% range
across all three deployment models,
whereas Tool-Calling drops from 30.0\% with gpt-oss-120b
to 6.7\% with Qwen3.5-4B.
The largest gain therefore appears
with the smallest deployment model.
This pattern suggests that reusable code
supplies recurring browser control
that smaller models would otherwise need
to reconstruct during each execution.

\subsection{Harness Growth Convergence Analysis}
\label{sec:results:harness-analysis}

\begin{figure}[t]
    \centering
    \includegraphics[
        width=\figwidth, trim=20pt 20pt 5pt 10pt, clip
    ]{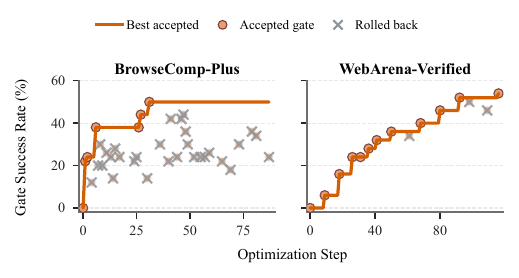}
    \caption{%
        Validation convergence
        (training steps \versus{} task success rate in \%)
        during harness training
        on BrowseComp-Plus and WebArena-Verified.
        Each point corresponds to a gate evaluation
        at the indicated optimization step.
    } \label{fig:gate-training-curves}
\end{figure}

\noindent\textbf{Shared failures become reusable control.}
The BrowseComp-Plus harness grows a single
retrieval and verification pipeline
that every task reuses for query generation,
evidence collection and compression,
and answer checking.
Repairs to this common path
therefore transfer across gate tasks
instead of adding instance-specific branches,
which explains the rapid early gain
in \Cref{fig:gate-training-curves}.

\noindent\textbf{Specialized handlers grow without regressions.}
WebArena-Verified improves across the training run
and rolls back only a few candidates.
Its learned program adds specialized handlers
for diverse Shopping, Reddit, and Map tasks
around a general LLM-guided browser loop.
These branches add new behavior
without rewriting paths used by other task types,
which is consistent with the steadier training curve.

\noindent\textbf{The task family shapes the controller.}
Both harnesses start from the same strategy-free seed program,
yet one becomes a shared pipeline
and the other a set of specialized handlers.
This contrast supports the low-prior-commitment goal:
harness growth lets executable structure emerge from task feedback
instead of fixing a controller in advance.
Because the benchmarks use different training configurations,
we treat the contrast as a description of these runs,
not as a comparison of intrinsic difficulty.

\subsection{Ablation Studies}\label{sec:results:ablation}

We isolate the three principal mechanisms
in \Cref{alg:harness-growth}:
function-level guidance,
the failure-window curriculum,
and held-out gate validation with transactional rollback.
All ablations are conducted on BrowseComp-Plus
using gpt-oss-20b
as the deployment model
and GPT-5.6-terra (High)
as the optimizer.
Each configuration is optimized once for 10 optimization steps
and evaluated on the same 50-task final-evaluation set.
\textsc{w/o Function-Level Guidance}
removes function-level traces,
trace-derived edit localization,
and the edit budget \(L\),
allowing unconstrained whole-program edits.
\textsc{w/o Gate Validation}
disables gate-based acceptance and transactional rollback.
\textsc{w/o Failure-Window}
sets the window capacity to \(K=1\),
so each repair is conditioned on a single active failure.
All remaining settings are held fixed.

\begin{figure}[t]
    \centering
    \includegraphics[
        width=\figwidth, trim=0 20pt 0 10pt, clip
    ]{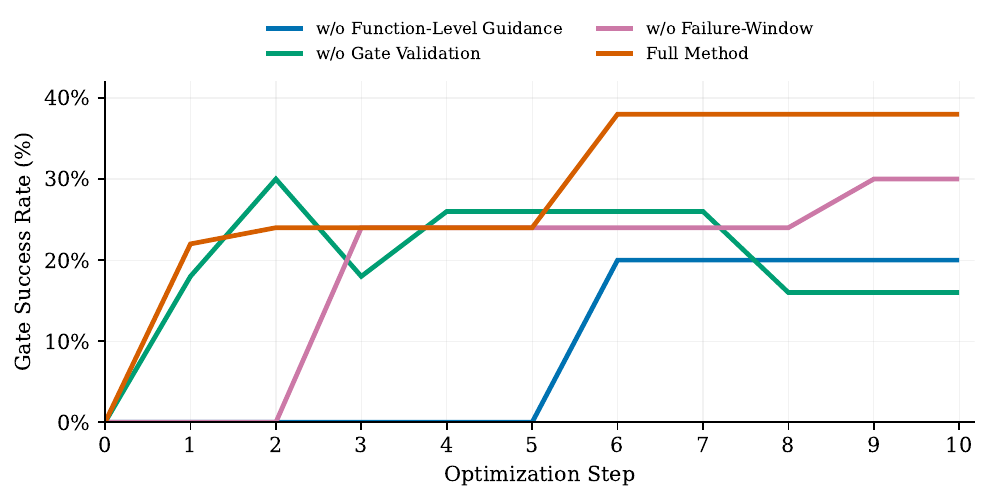}
    \caption{%
        Validation success rates over 10 optimization steps on BrowseComp-Plus.
        Each variant removes one mechanism from the full method.
        Each trajectory is obtained from a single optimization run.
    } \label{fig:ablation-gate-first10}
\end{figure}

Each removal lowers final-evaluation success,
but the optimization trajectories reveal
why the mechanisms complement one another
(\Cref{tab:ablation-results};
\Cref{fig:ablation-gate-first10}).
\begin{table}[t]
\centering
\setlength{\tabcolsep}{4pt}
\begin{adjustbox}{max width=\linewidth}
\begin{tabular}{lc}
\toprule
\textbf{Variant}
& \textbf{Success (\%) $\uparrow$} \\
\midrule
w/o Function-Level Guidance & 18.0 \\
w/o Gate Validation & 22.0 \\
w/o Failure-Window & 28.0 \\
Full Method & 36.0 \\
\bottomrule
\end{tabular}
\end{adjustbox}
\caption{Single-run ablation results on the 50-task BrowseComp-Plus
final-evaluation set. Each configuration is optimized for 10 steps.}
\label{tab:ablation-results}
\end{table}

\noindent\textbf{Trace locality makes program search effective.}
Removing function-level guidance
halves final success from 36\% to 18\%.
The whole-program variant also makes no gate progress
for five steps before reaching a lower plateau.
This stall supports the role of function-level traces
in linking a task failure to a bounded edit surface.

\noindent\textbf{Gate rollback is necessary.}
Without gate validation,
gate success rises to 30\%
but then falls to 16\%.
The full method instead retains its best gate result
once it reaches it.
The contrast directly shows the failure mode
that rollback targets:
a repair can solve current failures
while damaging behavior acquired earlier.

\noindent\textbf{Multi-failure windows favor general reuse.}
Conditioning each edit on one failure
does not cause the same stall or regression,
but it lowers final success by 8 pp.
This gap suggests that joint repair
helps the optimizer identify behavior shared across failures.
\FloatBarrier

\section{Conclusion}\label{sec:conclusion}

Agents that repeatedly serve one task family
need not reconstruct all control through LLM inference.
\method{} instead acquires recurring control as persistent code,
while keeping the LLM
for task-specific semantic decisions.
Starting from a strategy-free scaffold,
function-level traces focus edits
on code implicated by current failures,
a bounded failure window encourages repairs shared across tasks,
and a success-first held-out gate
rolls back changes that reduce prior capability.
The method does not prescribe a controller;
it lets executable structure adapt to task feedback.

Harness growth thus provides a structural way
to manage model context:
encode repeatable control in the harness
and reserve model context
for evidence and decisions that vary by task.
Its value depends on reusing the learned harness
enough to offset offline optimization.
Real deployment also requires sandboxing,
explicit permission boundaries,
and validation of generated code.

\FloatBarrier
\bibliographystyle{aaai2027}
\bibliography{references}

\clearpage
\appendix
\section{Experimental Details}\label{app:setup}

\subsection{Datasets}\label{app:setup:datasets}

For each benchmark, we partition tasks into train, gate, and final splits
with 200 train tasks, 50 gate tasks, and 50 final tasks.
The train split provides optimization traces, the gate split is used for checkpoint selection, and the final split is held out for reporting results.
For WebArena-Verified, we restrict the benchmark to tasks involving the
\texttt{shopping}, \texttt{reddit}, and \texttt{map} websites, and use
stratified sampling over intent templates, task types, and websites
to improve coverage and balance across task families, operation types,
and websites.
All dataset splits are constructed using a fixed random seed of 42.

\subsection{Shared Evaluation Configuration}
\label{app:setup:shared-evaluation}

All final-evaluation runs use the shared configuration in
\Cref{tab:shared-evaluation-config}.
The agent is not explicitly informed of the remaining call budget;
the executable harness enforces the limit.

\begin{table}[H]
\centering
\small
\begin{adjustbox}{max width=\linewidth}
\begin{tabular}{ll}
\toprule
\textbf{Parameter} & \textbf{Configuration} \\
\midrule
Sampling temperature
& 1.0 \\
Reasoning effort
& Medium \\
Maximum online LLM calls
& 50/task \\
Maximum output
& 8192 tokens/call \\
Task timeout
& 1,800 s \\
\bottomrule
\end{tabular}
\end{adjustbox}
\caption{Shared configuration used for final evaluation of all methods.}
\label{tab:shared-evaluation-config}
\end{table}

\subsection{Baseline Configurations}
\label{app:setup:baselines}

All baselines use the same seed interface,
tool APIs,
deployment models,
and LLM-call budget as \method{}.

\noindent\textbf{BrowseComp-Plus.}\\
All methods use the same search tool.
\textbf{Tool-Calling} alternates LLM reasoning with search calls until
the model emits a final answer.
\textbf{Self-Ask} decomposes the task into follow-up search questions
and intermediate answers before producing the final answer.
\textbf{IRCoT} interleaves retrieval with evidence-sentence generation
and then uses a final QA call.

\noindent\textbf{WebArena-Verified.}\\
All methods use text-only browser tools without
screenshots.
\textbf{Tool-Calling} emits one browser action or final JSON answer per
step.
\textbf{WebDreamer} proposes candidate browser actions and selects among
them using world-model and value-model scoring.
\textbf{AgentOccam} uses compressed observations and short interaction
history, with branch, prune, and note operations.
WebDreamer and AgentOccam are adapted to the seed interface,
accessibility-tree observation space,
unified LLM budget,
and shared deployment models.

\subsection{Token Prices}\label{app:setup:token-prices}

We compute online cost from the provider token prices in
\Cref{tab:model-prices} and exclude judge calls.

\begin{center}
\scriptsize
\setlength{\tabcolsep}{5pt}
\renewcommand{\arraystretch}{0.95}
\begin{adjustbox}{max width=\linewidth}
\begin{tabular}{llrrr}
\toprule
\textbf{Model} & \textbf{Provider}
& \textbf{Input} & \textbf{Cache Read} & \textbf{Output} \\
\midrule
gpt-oss-120b & Groq & \$0.15 & \$0.075 & \$0.60 \\
gpt-oss-20b & Groq & \$0.075 & \$0.0375 & \$0.30 \\
qwen3.5-4b & Fireworks & \$0.20 & \$0.10 & \$0.20 \\
\bottomrule
\end{tabular}
\end{adjustbox}
\vspace{-0.35em}
\captionof{table}{Token prices (USD per million tokens).}
\label{tab:model-prices}
\end{center}

\subsection{Hyperparameters}\label{app:setup:hyperparameters}

We select hyperparameters using only the training and gate splits;
\Cref{tab:training-hyperparameters} reports the fixed configurations
used for harness training.

\begin{table}[H]
\centering
\small
\begin{adjustbox}{max width=\linewidth}
\begin{tabular}{lcc}
\toprule
\textbf{Parameter}
& \textbf{BrowseComp-Plus}
& \textbf{WebArena-Verified} \\
\midrule
Training-time deployment LLM
& gpt-oss-20b & gpt-oss-120b \\
Optimizer
& GPT-5.6-terra & GPT-5.4 \\
Reasoning effort
& High & High \\
Deployment-LLM sampling seed
& 42 & 42 \\
Failure-window capacity \(K\)
& 8 & 4 \\
Per-task repair budget \(R_{\max}\)
& 5 & 5 \\
Gate interval \(Q\)
& 1 & 8 \\
Function-level edit budget \(L\)
& 10 & 10 \\
Candidates per optimization step
& 1 & 1 \\
Maximum online LLM calls
& 50/task & 50/task \\
Task timeout
& 900 s & 1,800 s \\
\bottomrule
\end{tabular}
\end{adjustbox}
\caption{Benchmark-specific hyperparameters used for harness training.
All configurations are selected without access to the final-evaluation
sets.}
\label{tab:training-hyperparameters}
\end{table}

\clearpage

\iftoggle{arxiv}{\begin{center}}{\begin{strip}}
\centering
\includegraphics[
    width=\iftoggle{arxiv}{\textwidth}{0.80\textwidth}
]{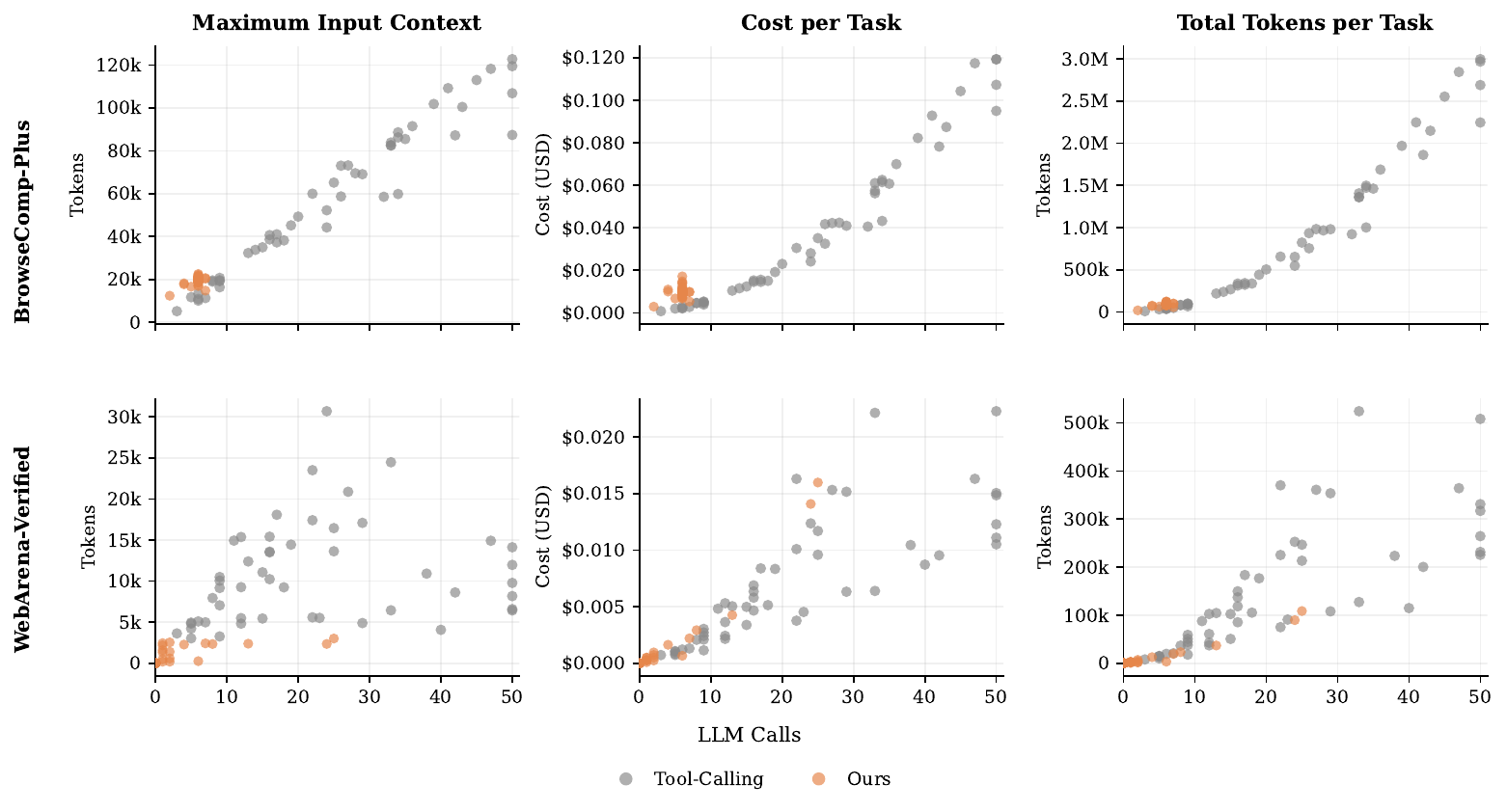}
\captionof{figure}{Task-level comparison between \method{} and Tool-Calling.
Each panel compares per-task LLM calls against one efficiency statistic
on BrowseComp-Plus and WebArena-Verified using GPT-OSS-20B.
Cost and token measurements exclude judge usage.}
\label{fig:app:ours-vs-tool-calling-six-panel}
\par\vspace{0.2em}
\centering
\begin{minipage}[t]{0.48\textwidth}
\centering
\begin{minipage}[c][0.45\textheight][c]{\linewidth}
\centering
\includegraphics[
    width=\linewidth,
    height=0.45\textheight,
    trim=0 0 0 4,
    clip,
    keepaspectratio
]{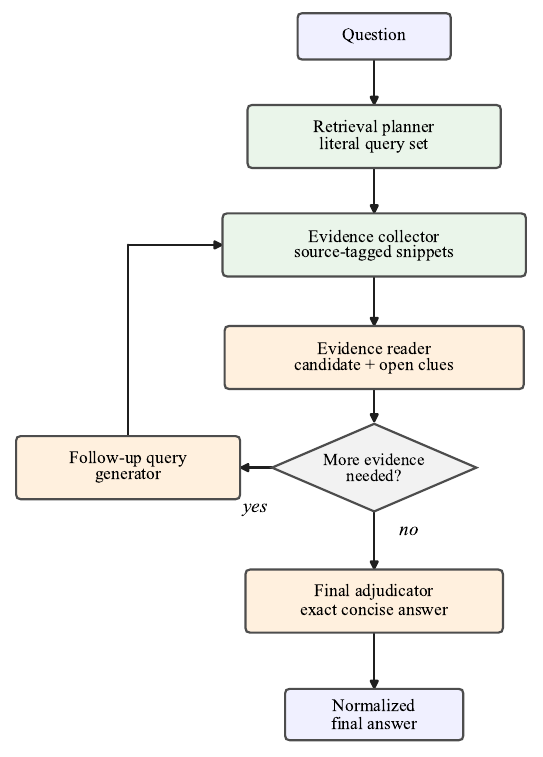}
\end{minipage}

\vspace{0.15em}
\small (a) BrowseComp-Plus.
\end{minipage}
\hfill
\begin{minipage}[t]{0.48\textwidth}
\centering
\begin{minipage}[c][0.45\textheight][c]{\linewidth}
\centering
\includegraphics[
    width=\linewidth,
    height=0.45\textheight,
    trim=0 0 0 4,
    clip,
    keepaspectratio
]{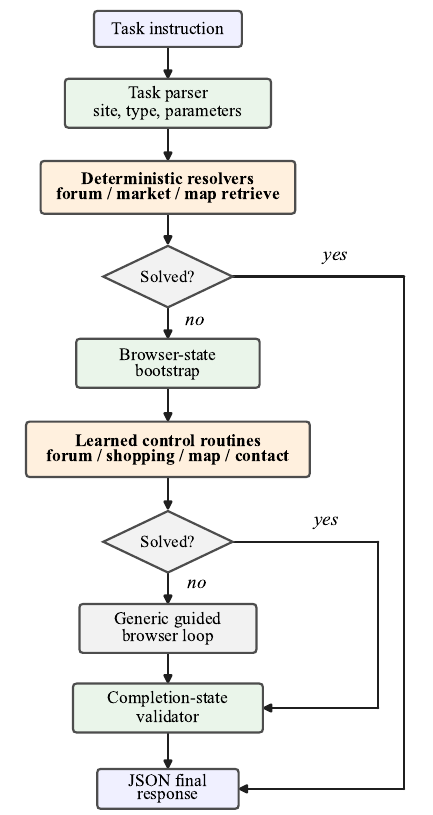}
\end{minipage}

\vspace{0.15em}
\small (b) WebArena-Verified.
\end{minipage}
\captionof{figure}{Post-hoc abstraction of the learned executable
harnesses \(h^\star\). BrowseComp-Plus induces a shared retrieval and
evidence-verification pipeline, while WebArena-Verified induces a
layered web-control program with deterministic resolvers, learned
control routines, fallback browser control, and completion validation.}
\label{fig:app:discovered-harnesses}
\iftoggle{arxiv}{\end{center}}{\end{strip}}

\section{Additional Results}
\label{app:results}

\Cref{fig:app:ours-vs-tool-calling-six-panel} provides task-level
efficiency measurements corresponding to the aggregate results in the
main paper.
The figure reports per-task calls, cost, and token usage for \method{}
and Tool-Calling under the same 50-call evaluation budget.

\subsection{Discovered Harness Structures}
\label{app:results:discovered-harnesses}

To make the learned programs interpretable,
\Cref{fig:app:discovered-harnesses}
abstracts the final harnesses \(h^\star\)
into their main executable control paths.
These diagrams are post-hoc summaries of the discovered code,
not hand-designed architectures used by the optimizer.
They illustrate how failure-guided harness growth
adapts the executable harness to the benchmark:
BrowseComp-Plus induces a shared retrieval and evidence-verification
pipeline,
whereas WebArena-Verified induces a controller that combines
deterministic resolvers, learned control routines,
a generic guided browser loop,
and an explicit completion-state validator.
In both cases, the harness implements reusable control around the same
fixed deployment LLM and tool interfaces.

\end{document}